\documentclass{article} 
\usepackage{iclr2026_conference,times}

\usepackage{amsmath,amsfonts,bm}

\def\eqref#1{equation~\ref{#1}}

\def\1{\bm{1}}

\DeclareMathAlphabet{\mathsfit}{\encodingdefault}{\sfdefault}{m}{sl}
\SetMathAlphabet{\mathsfit}{bold}{\encodingdefault}{\sfdefault}{bx}{n}

\usepackage{hyperref}
\usepackage{url}

\usepackage{algorithm}
\usepackage{algorithmic}
\usepackage{amsmath}
\usepackage{amssymb}
\usepackage{multirow}
\usepackage{booktabs}
\usepackage{newfloat}
\usepackage{listings}
\usepackage{graphicx} 
\usepackage{subcaption}
\usepackage{makecell}
\usepackage{wrapfig}

\title{ChartMaster: Advancing Chart-to-Code Generation with Real-World Charts and Chart Similarity Reinforcement Learning}

\author{Wentao Tan$^{1,2}$~~
    Qiong Cao$^{2}$\thanks{Corresponding author}~~
    Chao Xue$^2$~~
    Yibing Zhan$^{3}$~~
    Changxing Ding$^{1}$~~
    Xiaodong He$^{2}$~~ \\
$^1$South China University of Technology \quad 
    $^2$JD Future Academy, China \quad 
    $^3$Wuhan University \\
\texttt{ftwentaotan@mail.scut.edu.cn} \\ \texttt{\{caoqiong1, xuechao19,  xiaodong.he\}@jd.com} \\
\texttt{zybjy@mail.ustc.edu.cn \quad chxding@scut.edu.cn }
}

\iclrfinalcopy 
\begin{document}

\maketitle

\begin{abstract}

The chart-to-code generation task requires MLLMs to convert chart images into executable code. This task faces two main challenges: limited data diversity and the difficulty of maintaining visual consistency between generated charts and the original ones. Existing datasets mainly rely on synthetic seed data to prompt GPT models for code generation, resulting in homogeneous samples that limit model generalization to real-world chart styles. To address this, we propose \textbf{ReChartPrompt}, leveraging real-world, human-designed charts extracted from arXiv papers as prompts. By harnessing the rich content and diverse visual styles of arXiv charts, we construct ReChartPrompt-240K, a large-scale and highly diverse dataset that better reflects realistic chart variations.
For the second challenge, although SFT improves code understanding by optimizing next-token prediction, it does not provide direct supervision on visual features. As a result, it often fails to guarantee that the generated charts visually match the original ones. To address this, we propose \textbf{ChartSimRL}, a GRPO-based reinforcement learning algorithm guided by a novel chart similarity reward. This reward consists of two components: \textit{attribute similarity}, which measures the overlap of chart attributes like layout and color between the generated and original charts, and \textit{visual similarity}, which evaluates overall visual features, including texture, using convolutional neural networks. Unlike traditional text-based rewards, our reward accounts for the multimodal nature of the chart-to-code generation task, significantly enhancing the model's ability to accurately reproduce charts.
Integrating ReChartPrompt and ChartSimRL, we develop the \textbf{ChartMaster} model, achieving SOTA results among 7B-parameter models and rivaling GPT-4o on various chart-to-code benchmarks.
All resources are available at \url{https://github.com/WentaoTan/ChartMaster}.
\end{abstract}

\section{Introduction}

The chart-to-code generation task aims to automatically convert chart images into executable code \citep{yang2024chartmimic}, enabling applications including automated data analysis, report generation, and intelligent question answering \citep{zhao2025chartcoder}. This task is challenging as it requires accurate visual understanding, cross-modal reasoning, and advanced code synthesis. Although recent advances in Multimodal Large Language Models (MLLMs) show promising results in various vision-language tasks, their performance on chart-to-code generation remains limited due to the unique complexity of charts and the need for precise code output.

Prior work, such as ChartCoder \citep{zhao2025chartcoder}, advanced the field by building the large Chart2Code-160K dataset. This dataset is synthesized by guiding GPT-4o \citep{hurst2024gpt} with predefined chart attributes like chart type, color, and text. While this approach reduces the need for costly manual annotations and achieves strong performance, relying on predefined attribute seeds can introduce homogeneity and limit variability in the resulting dataset (see Appendix Fig. \ref{fig:dataset_visual}), potentially restricting model generalization to diverse real-world charts.

\begin{figure}[ht]
\centering
\setlength{\abovecaptionskip}{2pt}
\includegraphics[width=1.0\columnwidth]
{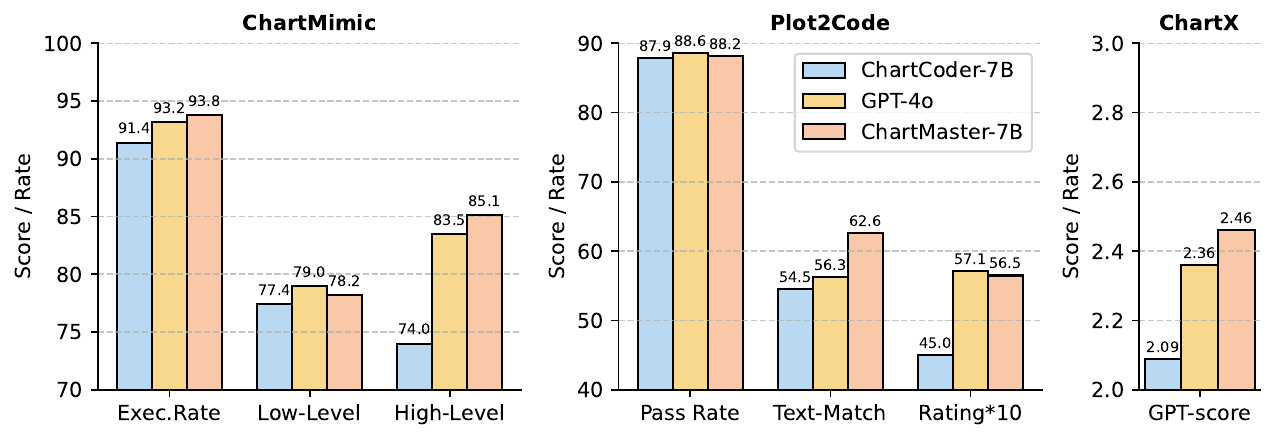}
\caption{Performance comparison on three benchmarks. Our method outperforms ChartCoder-7B \citep{zhao2025chartcoder}, and matches or exceeds GPT-4o on certain metrics. For better representation, the ``Rating" metric in the Plot2Code benchmark is multiplied by 10.
}
\label{fig1}
\vspace{-1em}
\end{figure}

To address this limitation, we introduce Real-world Chart Prompt Code Generation (ReChartPrompt), a novel automated pipeline that extracts real chart images from arXiv papers and leverages the Qwen2.5-VL-72B model \citep{bai2025qwen2} to generate corresponding code. 
By collecting 30,071 papers and utilizing their author-designed charts as prompts, we construct ReChartPrompt-240K, a large-scale dataset comprising 240K chart–code pairs. 
Since these charts originate from papers across diverse research fields and exhibit a wide variety of design styles, the dataset captures rich visual and semantic diversity, as illustrated in Fig.~\ref{fig:dataset_visual}. This heterogeneity enables effective generalization to real-world scenarios. 

While supervised fine-tuning with diverse data can help models generate better chart code, such next-token prediction alone does not ensure the output charts are visually faithful to the references. As shown in Fig.~\ref{fig:dingxing}, the SFT model produces charts closer to the ground truth than the baseline, but noticeable discrepancies remain in color, element positioning, and other visual attributes. To address this, we propose ChartSimRL, a reinforcement learning algorithm based on Group Relative Policy Optimization (GRPO) \citep{shao2024deepseekmath}, guided by a novel chart similarity reward. Specifically, the reward jointly considers (1) \textit{attribute similarity} that evaluates the consistency of chart elements such as textual content, numerical values, layout and color, and (2) \textit{visual similarity}, which assesses holistic visual resemblance using convolutional neural networks (e.g., ResNet \citep{he2016deep}) to extract and compare visual features. To the best of our knowledge, this is the first reward system that explicitly enforces multimodal visual-semantic consistency for chart-to-code generation. By encouraging models to produce code that renders charts both semantically accurate and visually faithful, we address a critical gap in prior research and support more robust generalization to real-world chart reproduction.

In summary, we introduce ChartMaster, an efficient framework for chart-to-code generation that combines the ReChartPrompt data generation pipeline with the ChartSimRL reinforcement learning strategy. Our key contributions are: (1) ReChartPrompt, an automated method for generating diverse datasets from real-world charts; (2) ChartSimRL, a reinforcement learning algorithm that uses both visual and attribute similarity to improve output; and (3) ChartMaster-7B, a compact model that delivers near GPT-4o performance with only 7 billion parameters. Fig.~\ref{fig1} highlights its efficiency and effectiveness.

\section{Related Work}

\subsection{Multimodal Code Generation}

Multimodal large language models (MLLMs) have recently demonstrated strong capabilities in code generation \citep{zhang2024humaneval,tan2025answers}. Notably, MMCode \citep{li2024mmcode} targets algorithmic problems embedded in visually rich contexts, where tasks are accompanied by one or more images.

Among multimodal code generation tasks, chart-to-code translation has emerged as a critical challenge \citep{yang2024matplotagent}. Existing benchmarks include Design2Code \citep{si2024design2code}, which evaluates HTML generation using CLIP scores \citep{radford2021learning} and structural HTML metrics, and Plot2Code \citep{wu2024plot2code}, which assesses both code correctness and visual fidelity. However, since the datasets for Design2Code and Plot2Code are sourced from the web, there is a risk of data leakage, which may compromise the reliability of model evaluation. To address this issue, ChartMimic \citep{yang2024chartmimic} provides a manually curated dataset of 4,800 chart-code pairs, along with additional fine-grained evaluation metrics.

Despite these benchmarks, large-scale chart-to-code training datasets remain scarce. ChartCoder \citep{zhao2025chartcoder} addresses this by creating Chart2Code-160K, the first large-scale training set generated by guiding GPT-4o with predefined chart attributes such as type, color, values, and titles. It further employs the ``Snippet of Thought” strategy \citep{zheng2023outline,luo2024python} to decompose code generation into structured steps, significantly boosting chart reasoning. Yet, reliance on fixed attributes limits chart diversity. In contrast, our ReChartPrompt leverages real-world charts from arXiv papers as prompts, yielding more diverse and representative chart–code pairs.

\subsection{Reinforcement Learning for MLLMs}
Reinforcement learning (RL) effectively enhances model capabilities \cite{wang2024comprehensive,milani2024explainable,tan2025beyond}. For example, RL from human feedback (RLHF) \cite{bai2022training} and direct preference optimization (DPO) \cite{rafailov2023direct} aligned model outputs with human preferences, improving complex reasoning and output quality.
Building on these advances, Group Relative Policy Optimization (GRPO) \cite{shao2024deepseekmath} was proposed as a novel RL algorithm that updated policies using relative rewards computed from groups of samples. DeepSeek-R1 \cite{guo2025deepseek} employed simple yet effective rewards based on output accuracy and response format, which enabled stable training and emergent reasoning such as reflection and ``a-ha” moments.

Inspired by DeepSeek-R1’s success, recent work extended GRPO-based RL to MLLMs \citep{tan2025reason,zhang2025r1,peng2025lmm,shen2025vlm} in two main directions. The first adapts R1’s method to MLLMs—for instance, Vision-R1 \citep{huang2025vision} uses SFT data with reflection for cold-start training and applies GRPO with accuracy- and format-based rewards. Similarly, MM-EUREKA \citep{meng2025mm} refines reward design and loss functions, successfully reproducing the visual “aha moment,” where the model revisits images “upon closer inspection.” These works primarily focus on mathematical reasoning tasks. The second direction applies GRPO to broader tasks such as visual perception \citep{yu2025perception}, segmentation \citep{liu2025seg}, and grounding \citep{zhang2025improving}, demonstrating its robustness and generalizability across domains.

However, to our knowledge, GRPO has not been applied to chart-to-code generation, mainly due to the challenge of designing reward functions that encourage generated code to faithfully reproduce charts both semantically and visually. We address this by proposing a novel chart similarity reward, significantly improving chart reproduction quality.

\section{Method}
\begin{figure*}[t]
\setlength{\abovecaptionskip}{2pt}
\centering
\includegraphics[width=1.0\linewidth]{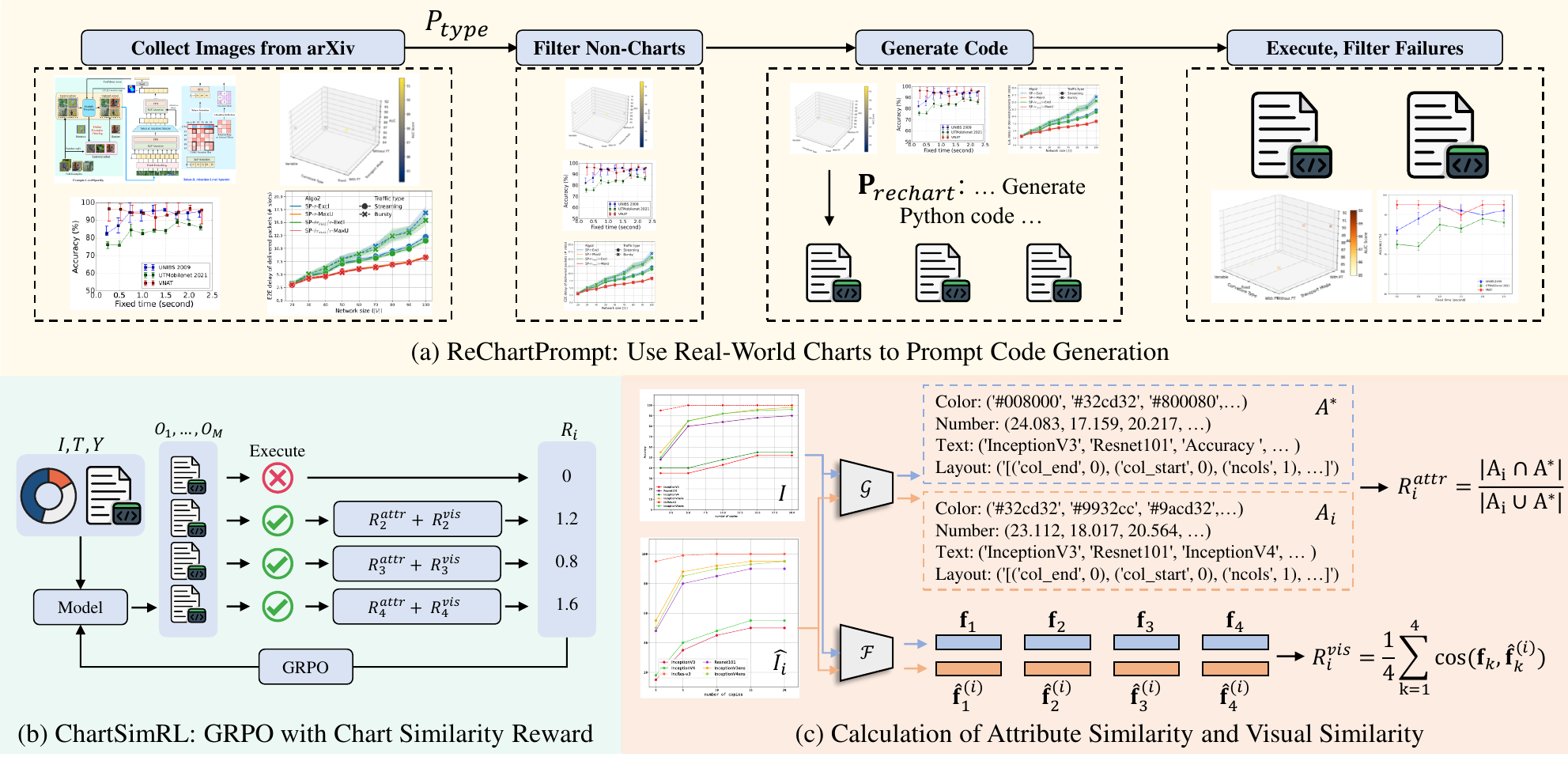} %
\caption{The overall framework of ChartMaster. (a) Real-world charts from arXiv are curated to create the ReChartPrompt-240K dataset for SFT (SFT is omitted in the figure). (b) The model is further optimized with ChartSimRL. (c) The definition of Chart Similarity: $\mathcal{G}$ denotes the semantic attribute extraction tool; $\mathcal{F}$ is the CNN-based feature extractor; and $\mathbf{f}$ is the extracted feature vector.}

\label{fig_overall}
\vspace{-1em}
\end{figure*}

Fig. \ref{fig_overall} illustrates the overall framework of ChartMaster, which consists of two main stages: data generation and model training.

\subsection{Using Real-World Charts to Generate Dataset}

To improve dataset diversity, we use real-world chart images as input to guide code generation, as shown in Fig. \ref{fig_overall} (a). This approach captures richer styles and content that predefined attribute seeds cannot represent.

\textbf{(1) Collecting Images from arXiv.}  
We leverage the arXiv API and Python's \texttt{requests} library to download paper source files, including LaTeX sources and image files (\texttt{.pdf}, \texttt{.png}, \texttt{.jpg}). To ensure diversity, we query source files related to top conferences (e.g., ICLR) and journals (e.g., TPAMI), extracting all images for subsequent processing.

\textbf{(2) Filtering Non-Chart Images.}  
Since extracted images include various diagrams beyond charts, we use the Qwen2.5-VL-72B model to classify images into 12 predefined chart categories. Images outside these categories are discarded. Classification is performed by prompting the model with \(P_{\text{type}}\) (see Fig. \ref{fig:prompt_type} in Appendix) to assign chart types.

\textbf{(3) Generating code with ReChartPrompt.}
The Qwen2.5-VL-72B model has demonstrated strong chart-to-code generation capabilities. As an open-source model, it is easily deployed via the vLLM framework \citep{kwon2023efficient}, making it well-suited for large-scale data generation. We design a set of 20 chart-to-code prompts to enrich instruction diversity, collectively referred to as \(\mathbf{P}_{\text{rechart}}\) (see Fig. \ref{fig:prompt_rechart} in Appendix). Below is an example:
\textit{$<$Real-World Chart$>$Please generate Python matplotlib code to recreate the picture shown.}

\textbf{(4) Code Execution, Filtering, and Dataset Construction.}
Generated code snippets may suffer from two issues: (a) execution errors caused by non-existent packages or syntax mistakes, and (b) discrepancies between the generated charts and the original images. To mitigate these problems, we execute all generated code and discard those that fail at runtime. We then pair the successfully executed code outputs with their generated images and instructions to form the final training triplets.

\textbf{Summary.}  
We download 30,071 papers from arXiv and extract their figures, filtering out non-chart ones to obtain 288,992 chart images. Using these charts, the Qwen2.5-VL-72B model generates corresponding code. After executing the generated code and removing failed cases, we collect 242,479 high-quality triplets that constitute the \textbf{ReChartPrompt-240K} dataset. Formally, the dataset is defined as $\mathcal{D} = \{(I_i, T_i, Y_i)\}_{i=1}^{N}$, where \(I_i\) represents a chart image, \(T_i \in \mathbf{P}_{\text{rechart}}\) is the instruction prompt, and \(Y_i\) denotes the executable code. Notably, all real-world chart data and generation models employed in this process are open-source, ensuring minimal cost and excellent scalability.

\subsection{Training ChartMaster: SFT and ChartSimRL}

ChartMaster is trained in two stages: (1) SFT on the ReChartPrompt-240K dataset to establish a solid foundation; and (2) further optimized with ChartSimRL to address the limitations of SFT's next-token prediction in maintaining visual consistency.

\textbf{Supervised Fine-Tuning.}
We conduct SFT by maximizing the likelihood of ground-truth code \(Y_i\) given chart image \(I_i\) and instruction \(T_i\):
\[
J_{\mathrm{SFT}}(\theta) = -\frac{1}{N} \sum_{i=1}^N \log \pi_{\theta}(Y_i \mid I_i, T_i).
\]
\textbf{Reinforcement Learning with ChartSimRL.}
While SFT strengthens the model's basic capability, discrepancies may still exist between the generated charts and the originals (see Fig. \ref{fig:dingxing}). To further improve reproduction fidelity, we continue training the model using ChartSimRL, as illustrated in Fig. \ref{fig_overall} (b).
Specifically, for each training sample \((I, T, Y)\), the model samples a group of \(M\) candidate codes:
\[
\{O_1, O_2, \dots, O_M\} \sim \pi_{\theta}(\cdot \mid I, T).
\]
Each candidate code $O_i$ is then executed to generate a chart image 
$\hat{I}_i$, which is subsequently compared with the original chart 
${I}_i$ to compute a Chart Similarity Reward. If the execution of 
$O_i$ fails, the corresponding reward is set to zero.

\textbf{Chart Similarity Reward.}  
Traditional reward functions, such as the accuracy reward used in \citep{guo2025deepseek,huang2025vision}, primarily assess the consistency between generated text and ground-truth text. However, the chart-to-code task is inherently multimodal, involving both code and generated charts, requiring evaluation of not only semantic correctness but also visual alignment. To this end, we design a novel chart similarity reward as:
\[
R_i = R_i^{\mathrm{attr}} + R_i^{\mathrm{vis}}.
\]
Here, \(R_i^{\mathrm{attr}}\) measures the semantic consistency, and \(R_i^{\mathrm{vis}}\) captures visual similarity (see Fig. \ref{fig_overall} (c)).

\textit{Attribute Similarity:}  
We develop a semantic attribute extraction tool, denoted \(\mathcal{G}(\cdot)\), to obtain attribute sets from chart images and their code. Given \(\mathcal{A}_i = \mathcal{G}(\hat{I}_i, O_i)\) and \(\mathcal{A}^* = \mathcal{G}(I, Y)\), the semantic similarity \(R_i^{\mathrm{attr}}\) is computed as their Jaccard similarity:
\[
R_i^{\mathrm{attr}} = \frac{|\mathcal{A}_i \cap \mathcal{A}^*|}{|\mathcal{A}_i \cup \mathcal{A}^*|} \in [0,1].
\]
By design, \(R_i^{\mathrm{attr}} = 1\) indicates a perfect match of semantic attributes, while lower values reflect semantic discrepancies. To accommodate minor numerical variations, we consider numerical values \(a \in \mathcal{A}_i\) and \(b \in \mathcal{A}^*\) matching if \(|a - b| \leq 0.01 \times |b|\).

\textit{Visual Similarity:}
We use a pretrained ResNet-18 network \citep{he2016deep} \(\mathcal{F} = \{\mathcal{F}_1, \mathcal{F}_2, \mathcal{F}_3, \mathcal{F}_4\}\) to extract feature maps from both \(I\) and \(\hat{I}_i\). Here, \(\mathcal{F}_k(\cdot) \in \mathbb{R}^{C_k \times H_k \times W_k}\) denotes the output feature map of the \(k\)-th residual block. We extract the feature map and flatten them into vectors like:
\[
\begin{aligned}
& F_{k} = \mathcal{F}_{k}(I), \quad \quad && \hat{F}_{k}^{(i)} = \mathcal{F}_{k}(\hat{I}_i), \\
& \mathbf{f}_{k} = \mathrm{vec}(F_{k}) \in \mathbb{R}^{d_k}, \quad \quad && \hat{\mathbf{f}}_{k}^{(i)} = \mathrm{vec}(\hat{F}_{k}^{(i)}) \in \mathbb{R}^{d_k},
\end{aligned}
\]
where \(d_k = C_k \times H_k \times W_k\). The visual similarity reward is defined as the average cosine similarity between the corresponding feature vectors,
\[
R_i^{\mathrm{vis}} = \frac{1}{4} \sum_{k=1}^4 \frac{\mathbf{f}_{k} \cdot \hat{\mathbf{f}}_{k}^{(i)}}{\|\mathbf{f}_{k}\| \, \|\hat{\mathbf{f}}_{k}^{(i)}\|} \in [0,1].
\]

\textbf{Chart Similarity Reinforcement Learning.} 
We normalize rewards within a group of \(M\) candidates to compute relative advantages:
\[
\hat{A}_i = \frac{R_i - \mathrm{mean}(\{R_j\}_{j=1}^M)}{\mathrm{std}(\{R_j\}_{j=1}^M)},
\]
where \(\mathrm{mean}(\cdot)\) and \(\mathrm{std}(\cdot)\) denote the sample mean and standard deviation, respectively.

Following the GRPO framework~\citep{shao2024deepseekmath}, we update the model by maximizing the clipped surrogate objective with a KL penalty to stabilize training:
\begin{align}
J_{\mathrm{ChartSimRL}}(\theta) & = \mathbb{E}_{(I,T) \sim p_{\mathcal{D}}, \, \{o_i\}_{i=1}^M \sim \pi_{\mathrm{old}}(\cdot|I,T)} \nonumber \Bigg[
\frac{1}{M} \sum_{i=1}^M \min \Big(
 \frac{\pi_\theta(o_i | I,T)}{\pi_{\mathrm{old}}(o_i | I,T)} \hat{A}_i, \nonumber \\
& \operatorname{clip}\Big(\frac{\pi_\theta(o_i | I,T)}{\pi_{\mathrm{old}}(o_i | I,T)}, 1-\epsilon, 1+\epsilon\Big) \hat{A}_i
\Big) \nonumber - \beta D_{\mathrm{KL}}\big(\pi_\theta(\cdot|I,T) \| \pi_{\mathrm{ref}}(\cdot|I,T)\big)
\Bigg],
\label{eq:chartsimrl_obj}
\end{align}
where \(\pi_{\mathrm{old}}\) is the previous policy, \(\pi_{\mathrm{ref}}\) is the reference policy, \(\epsilon\) is the clipping hyperparameter, and \(\beta\) controls the KL regularization strength.

ChartSimRL guides the model to generate chart code that better aligns with the original charts' semantic and visual properties, significantly improving chart-to-code generation performance beyond what is achievable by supervised fine-tuning alone.

\textbf{Summary.} ReChartPrompt and ChartSimRL have been effectively integrated into the ChartMaster framework. This framework not only leverages real-world data for enhanced data diversity but also employs a novel algorithm to ensure visual and semantic alignment in chart reproduction. Consequently, ChartMaster stands as a comprehensive solution for the chart-to-code generation task, demonstrating marked improvements in performance and generalization capabilities.

\section{Experiment}

\begin{table*}[t]
\centering
\caption{Evaluation results of various MLLMs. Reported results are taken from existing benchmarks when available; missing results are supplemented by our own experiments. Among open-source 7B-scale models, our method achieves the best performance.}
\vspace{-1em}
\setlength{\tabcolsep}{1.0mm}
\resizebox{1.0\linewidth}{!}{
\begin{tabular}{l|ccc|ccc|c}
\toprule[1pt]
\multirow{2}{*}{Model} &  \multicolumn{3}{c|}{ChartMimic} & \multicolumn{3}{c|}{Plot2Code} & ChartX \\

   & Exec.Rate & Low-Level & High-Level & Pass Rate  & Text-Match  &Rating & GPT-score\\ \hline

Full score  & 100 & 100 & 100 & 100 & 100 &10 & 5 \\

\midrule
\multicolumn{8}{c}{\it{Closed-Source Model}} \\ 
\midrule
GeminiProVision \citep{team2023gemini}  & 68.2 & 53.8 & 53.3  & 68.2 & 53.6& 3.69 &-\\
Claude-3-opus \citep{anthropic2024claude}  & 83.3 & 60.5 & 60.1 & 84.1 & 57.5 & 3.80 & -\\
GPT-4V \citep{hurst2024gpt} & 91.2 & 76.4 & 78.9 & 84.1 & 57.7 & 5.58 & 2.63\\
GPT-4o \citep{hurst2024gpt} & 93.2 & 79.0 & 83.5 & 88.6 & 56.3 & 5.71 & 2.36\\

\midrule
\multicolumn{8}{c}{\it{Open-Source Model}} \\
\midrule
ChartAssisstant-13B \citep{meng2024chartassisstant}  & - & - &-  &-&-&-& 0.82 \\
ChartVLM-L-14B \citep{xia2024chartx}  & 19.5 & 15.8 & 13.9 & - & - & - &  1.58\\
DeepSeek-VL-7B \citep{lu2024deepseek}  & 41.3 & 19.0 & 17.6 & 64.4 & 32.6& 2.26 & -\\
TinyChart-3B \citep{zhang2024tinychart} &  42.5 & 26.3 & 25.9 & 43.2 & 44.6 & 2.19 & 1.89\\
ChartLlama-13B \citep{han2023chartllama} & 57.5 & 24.8 & 28.1 & 58.4 & 40.3 & 2.32& 0.94 \\
LLaVA-Next-Mistral-7B \citep{li2024llava}  & 59.7 & 20.7 & 21.3 & 72.0 & 38.7 & 2.87  &- \\
InternVL2-8B \citep{chen2024internvl}  & 61.8 & 34.4 & 38.9 & 77.3 & 37.1 & 2.78 & 1.63\\
Qwen2-VL-7B \citep{wang2024qwen2}  & 67.0 & 32.9 & 35.0 & 68.2 & 33.8 & 3.10 & 1.50\\
MiniCPM-Llama3-V2.5-8B \citep{yao2024minicpm}  & 80.3 & 36.6 & 42.1  & 76.3 & 37.3& 2.61&1.66\\
Qwen2-VL-72B \citep{wang2024qwen2}  & 73.3 &54.4  & 50.9 & 72.0 & {53.4}  & {4.26}  & 1.69  \\
InternVL2-Llama3-76B \citep{chen2024internvl} & 83.2 & 54.8 & 62.2  & 85.6 & 46.6 & 3.89&1.74 \\	
Qwen2.5-VL-72B \citep{bai2025qwen2}  & 88.5 & 72.7 & \underline{79.1} & 84.8 & \textbf{68.4} &\textbf{6.83}  &\textbf{2.52}  \\ 
ChartCoder-7B \citep{zhao2025chartcoder} & \underline{91.4} & \underline{77.4} & 74.0 & \underline{87.9} & 54.5 & 4.50 & 2.09 \\
\hline 
Qwen2.5-VL-7B (Baseline) \citep{bai2025qwen2} & 65.5 & 39.9 & 40.7 & 67.4 & 43.8 & 4.60 &  2.18 \\ 
ChartMaster-7B  & \textbf{93.8} & \textbf{78.2} & \textbf{85.1} &\textbf{88.2}  &\underline{62.6}  &\underline{5.65} & \underline{2.46} \\
\hline
\end{tabular}}
\label{tab:main_results_direct}
\end{table*}

\subsection{Comparison with SOTA}
We instantiate ChartMaster on the Qwen2.5-VL-7B backbone, resulting in the ChartMaster-7B model, and conduct comprehensive comparisons with a range of MLLMs. The detailed implementation and evaluation protocols are provided in the Appendix \ref{sec:details}.
As shown in Table~\ref{tab:main_results_direct}, ChartMaster-7B achieves state-of-the-art performance among open-source models at the 7B scale, showing competitive performance against GPT-4o. Notably, ChartMaster-7B consistently outperforms the baseline Qwen2.5-VL-7B across all metrics; for instance, in the ChartMimic benchmark, it improves both low-level and high-level metrics by about 40 percentage points. Furthermore, although our training dataset is derived from the larger Qwen2.5-VL-72B model—essentially a distillation-like setting—ChartMaster-7B still surpasses Qwen2.5-VL-72B on several benchmarks. These results convincingly demonstrate the effectiveness of the ChartMaster framework.

\subsection{Ablation Study}
\textbf{Ablation study on ChartMaster.} To assess the contribution of each component, we conduct an ablation study as summarized in Table~\ref{tab:ablation}. The base Qwen2.5-VL-7B model, without ReChartPrompt or ChartSimRL, demonstrates limited performance across benchmarks, revealing its restricted ability in both code generation and visual/semantic understanding. SFT with the ReChartPrompt-240K dataset leads to significant improvements in all metrics, demonstrating the high quality and effectiveness of ReChartPrompt-240K for chart-to-code generation. Further applying ChartSimRL on top of ReChartPrompt brings consistent gains, demonstrating its effectiveness in guiding the model to generate code that more accurately captures the original charts both semantically and visually. These findings confirm the complementary strengths of ReChartPrompt and ChartSimRL in enhancing ChartMaster’s overall performance.

\begin{table*}[t]
\centering
\caption{Ablation study on the contribution of each key component.}
\vspace{-1em}
\setlength{\tabcolsep}{1.0mm}
\resizebox{1.0\linewidth}{!}{
\begin{tabular}{cc||ccc|ccc|c}
\toprule[1pt]
\multirow{2}{*}{ReChartPrompt} & \multirow{2}{*}{ChartSimRL} & \multicolumn{3}{c|}{ChartMimic} & \multicolumn{3}{c|}{Plot2Code} & ChartX \\
  & & Exec.Rate & Low-Level & High-Level & Pass Rate  & Text-Match  &Rating & GPT-score\\ \hline
  & & 65.5 & 39.9 & 40.7 & 67.4 & 43.8 & 4.60 & 2.18\\
$\checkmark$ & & 91.1 & 73.7 & 80.9 & 80.3 & 59.3 & 5.34 & 2.36\\
$\checkmark$ & $\checkmark$& 93.8 & 78.2 & 85.1 &88.2  &62.6  &5.65  & 2.46\\

\hline
\end{tabular}}
\label{tab:ablation}
\vspace{-1em}
\end{table*}

\begin{table}[t]
  \centering
  \setlength{\tabcolsep}{1.0mm}
  \begin{minipage}{0.45\textwidth}
    \centering
    \caption{Ablation study of the Attribute and Visual similarity components in ChartSimRL.}
    \vspace{-1em}
    \resizebox{0.95\linewidth}{!}{
    \begin{tabular}{cc||ccc}
    \toprule[1pt]
      \multirow{2}{*}{$R_i^{\mathrm{attr}}$} & \multirow{2}{*}{$R_i^{\mathrm{vis}}$} & \multicolumn{3}{c}{ChartMimic} \\ 
       &  & Exec.Rate & Low-Level & High-Level \\ \hline
       &  & 91.1 & 73.7 & 80.9  \\ 
     $\checkmark$ & & 92.1 & 76.2 &  83.9 \\
      &$\checkmark$ & 92.1 & 77.7 &  84.3 \\
     $\checkmark$ & $\checkmark$ & 93.8 & 78.2 & 85.1  \\ \hline
    \end{tabular}}
    
    \label{tab:chartsim}
  \end{minipage}%
  \hfill
  \begin{minipage}{0.52\textwidth}
    \centering
    \caption{Ablation study of different attribute similarity metrics on the ChartMimic benchmark.}
    \vspace{-1em}
    \resizebox{1.0\linewidth}{!}{
    \begin{tabular}{c|c||ccc}
\toprule[1pt]
\multirow{2}{*}{$R_i^{\mathrm{attr}}$} & \multirow{2}{*}{Formula} &  \multicolumn{3}{c}{ChartMimic} \\
& & Exec.Rate & Low-Level & High-Level \\ \hline
- & - & 91.1 & 73.7 & 80.9  \\ \hline
Precision & $\frac{|\mathcal{A}_i \cap \mathcal{A}^*|}{|\mathcal{A}_i|}$ & 90.0  &72.6 & 79.0  \\
Recall & $\frac{|\mathcal{A}_i \cap \mathcal{A}^*|}{|\mathcal{A}^*|}$&90.6  &74.7  & 81.7  \\
F1 & $\frac{|\mathcal{A}_i \cap \mathcal{A}^*|}{(|\mathcal{A}_i| + |\mathcal{A}^*|)/2}$ &91.6 & 75.4 &\textbf{84.5}   \\
Jaccard & $\frac{|\mathcal{A}_i \cap \mathcal{A}^*|}{|\mathcal{A}_i \cup \mathcal{A}^*|}$ &\textbf{92.1} & \textbf{76.2} &  83.9 \\ \hline
\end{tabular}}
\label{tab:attr_sim}
  \end{minipage}
  \vspace{-1em}
\end{table}

\textbf{Ablation study on ChartSimRL.}
ChartSimRL introduces a novel multimodal chart similarity reward that combines both semantic similarity ($R_i^{\mathrm{attr}}$) and visual similarity ($R_i^{\mathrm{vis}}$) between the candidate and original charts. To dissect the contribution of each component, we conduct ablation experiments summarized in Table~\ref{tab:chartsim}. The results show that employing either $R_i^{\mathrm{attr}}$ or $R_i^{\mathrm{vis}}$ alone consistently improves performance across all evaluated metrics. Notably, the visual similarity reward yields more substantial gains, underscoring the critical importance of preserving visual fidelity in chart-to-code generation. Moreover, combining both rewards achieves the best overall results, demonstrating the advantage of a multi-faceted reward design that simultaneously captures semantic and visual aspects.

\textbf{Ablation study on Attribute Similarity.}
We adopt Jaccard similarity as a stringent metric for attribute similarity, whereby a candidate table achieves a perfect score only if its attribute set exactly matches that of the ground truth; even minor discrepancies incur penalties. To thoroughly assess the impact of different attribute similarity measures—Precision, Recall, F1 score, and Jaccard similarity—we conduct experiments on the ChartMimic benchmark, with results summarized in Table~\ref{tab:attr_sim}.

Our findings indicate that optimizing exclusively for Precision may lead to a slight decline in overall performance, as the model can achieve high Precision by predicting a limited subset of correct attributes while neglecting overall coverage. In contrast, Recall emphasizes coverage, which helps mitigate this issue and yields modest improvements. The F1 score, by harmoniously balancing Precision and Recall, further alleviates extreme biases and delivers enhanced overall performance. Notably, Jaccard similarity, measuring the intersection over union between predicted and reference attribute sets, enforces stricter penalties on both missing and redundant attributes. This higher overlap requirement enables Jaccard similarity to more faithfully capture the true semantic similarity between attribute sets, thereby resulting in the best overall performance.

\textbf{Ablation Study on Visual Similarity.}
We use ResNet-18 \citep{he2016deep} to extract features from charts to compute visual similarity. In fact, there are numerous methods to measure the similarity between two charts. To investigate the impact of different visual similarity metrics on model performance, we conduct an ablation study summarized in Table~\ref{tab:visual_sim}. Standard metrics such as MSE, SSIM \citep{wang2004image}, and PSNR \citep{hore2010image} primarily evaluate pixel-level or structural similarity (details in Appendix \ref{sec:standard_metrics}). The table shows that these metrics generally perform worse than more advanced methods. Notably, SSIM exhibits a significant decline in performance, indicating that pixel-based measures struggle to capture the complex visual nuances necessary for effective chart-to-code generation.

In contrast, CNN-based metrics like AlexNet \citep{krizhevsky2012imagenet}, VGG \citep{simonyan2014very}, and ResNet \citep{he2016deep}, which compare features in a learned representation space, consistently outperform both the baseline and pixel-level metrics across all evaluation criteria. Among them, ResNet-18 achieves the highest performance, highlighting the effectiveness of deep visual features.

For MLLM-based metrics, we leverage the high-level similarity prompt from ChartMimic combined with Qwen-2.5-VL-72B to evaluate the similarity between generated and reference charts. These metrics show improvements over the baseline. However, they still fall short of the best CNN-based metrics, suggesting that although MLLMs possess strong semantic understanding, further optimization is required for specialized visual tasks such as chart-to-code generation.

\begin{wraptable}{r}{0.5\textwidth}
\vspace{-1.5em}
\centering
\caption{Ablation study of different visual similarity metrics on the ChartMimic benchmark.}
\vspace{-1em}
\setlength{\tabcolsep}{1.0mm}
\resizebox{0.95\linewidth}{!}{
\begin{tabular}{c|ccc}
\toprule[1pt]
\multirow{2}{*}{$R_i^{\mathrm{vis}}$} & \multicolumn{3}{c}{ChartMimic} \\
 & Exec.Rate & Low-Level & High-Level \\ \hline
 - & 91.1 & 73.7 & 80.9  \\
 \midrule
\multicolumn{4}{l}{\textit{Standard Metrics:}} \\
\midrule
MSE & 91.1 & 73.6 & 77.9 \\
SSIM & 82.5 & 65.2 & 74.6  \\
PSNR & 91.4 & 75.1 & 82.1 \\
 \midrule
\multicolumn{4}{l}{\textit{CNN-Based Metrics:}} \\
\midrule
AlexNet & 90.3 & 74.7 & 82.6 \\
VGG  & 91.3 & 75.5 & 83.3 \\
ResNet-18  & \textbf{92.1} & \textbf{77.7} & \textbf{84.3} \\
 \midrule
\multicolumn{4}{l}{\textit{MLLM-Based Metrics:}} \\
\midrule
Qwen-2.5-VL-72B & 91.7 & 77.5 & 83.9 \\ \hline
\end{tabular}}
\label{tab:visual_sim}
\vspace{-1em}
\end{wraptable}

\begin{figure}[t]
\centering
\setlength{\abovecaptionskip}{2pt}
\includegraphics[width=1.0\columnwidth]{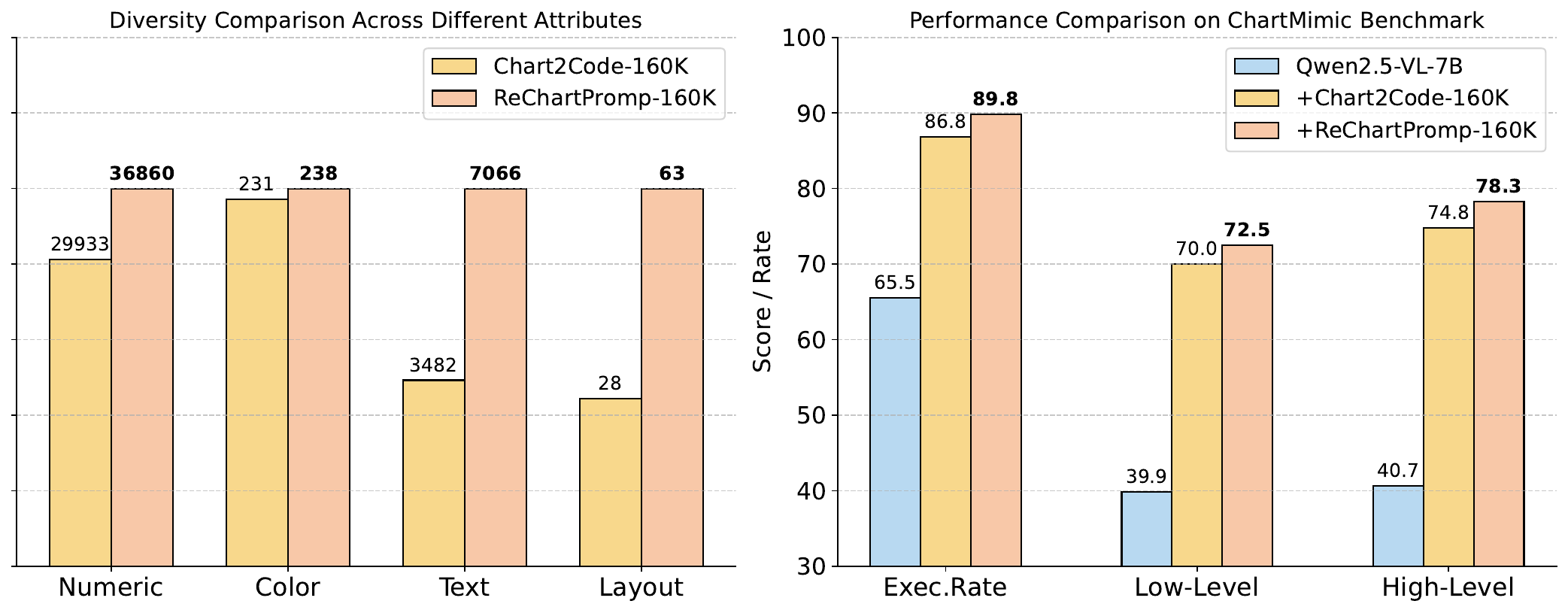} 
\caption{Comparison of diversity and fine-tuning results between Chart2Code-160K and ReChartPrompt-160K datasets.
}
\label{fig:advance_compare}
\vspace{-1em}
\end{figure}

\textbf{Comparison with Advanced Dataset.}
To comprehensively evaluate the diversity and quality of our dataset, we compare it with Chart2Code-160K \citep{zhao2025chartcoder}. For a fair comparison, we randomly sample 160K instances from our full dataset to construct the ReChartPromp-160K subset. Using the attribute extraction tool $\mathcal{G}(\cdot)$, we count unique chart attributes—including numerical values, colors, textual elements, and layouts—in both datasets. A higher number of unique attributes indicates greater attribute diversity.
As shown in the left panel of Fig.~\ref{fig:advance_compare}, ReChartPromp-160K exhibits a substantially richer attribute distribution across all categories, notably in text and layout. This advantage stems primarily from Chart2Code-160K's reliance on seed data sources, which results in repeated attribute patterns, whereas ReChartPromp-160K samples from distinct arXiv papers, ensuring broader coverage and less redundancy (see Appendix Fig.~\ref{fig:dataset_visual}).
This higher diversity brings clear benefits: as illustrated in the right panel, models fine-tuned on ReChartPromp-160K consistently outperform those trained on Chart2Code-160K, demonstrating the importance of attribute diversity for robust and effective chart understanding and code generation.

\subsection{Qualitative analysis}

\begin{figure}[t]
\centering
\setlength{\abovecaptionskip}{2pt}
\includegraphics[width=1.0\columnwidth]{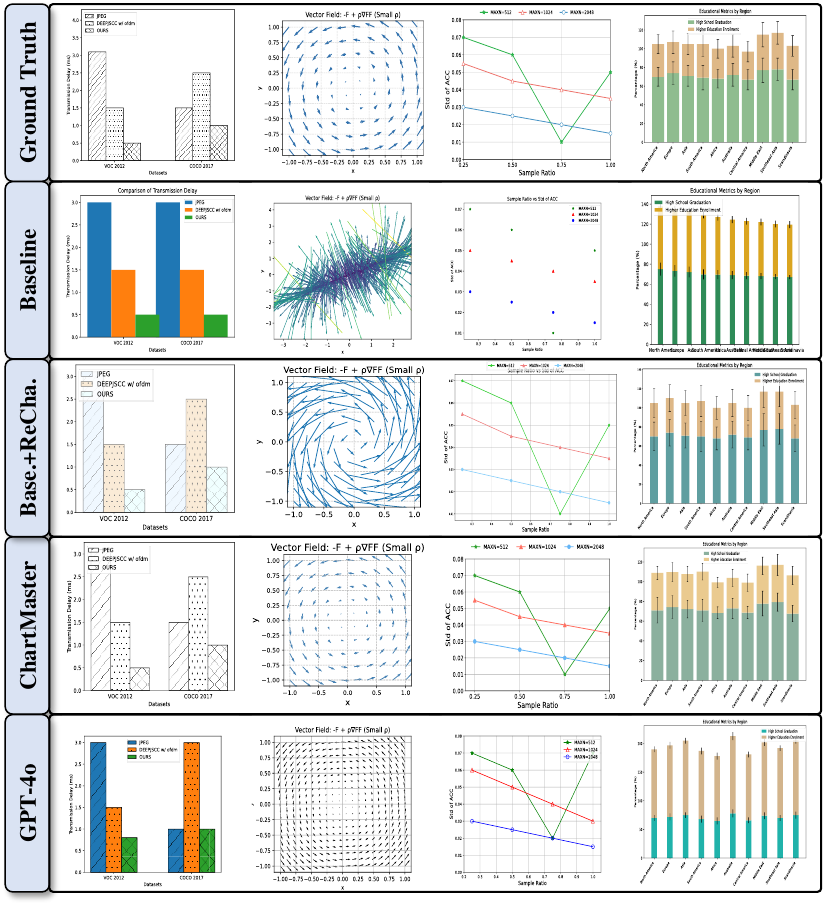} 
\caption{The test results of various models on the ChartMimic benchmark. ``Base.+ReCha." refers to the baseline model fine-tuned with the ReChartPrompt-240K dataset. Incorporating ReChartPrompt significantly enhances the chart-to-code generation capability of the base model, while ChartSimRL further improves the handling of fine details.
}
\label{fig:dingxing}
\vspace{-1em}
\end{figure}

Based on extensive experiments, we observe that ReChartPrompt generates charts with diverse and rich attributes, enabling the construction of a high-quality dataset that substantially enhances model performance. Building upon the distinctive features of the chart-to-code generation task, we propose the ChartSimRL algorithm, which further enhances the model’s capabilities. To comprehensively analyze the improvements brought by these contributions, we conduct a qualitative comparison of generated charts at different training stages on the ChartMimic benchmark (Fig.~\ref{fig:dingxing}). Our key findings are summarized as follows:
\textbf{(1) The baseline model produces basic chart layouts but often fails to replicate fine-grained visual details,} leading to noticeable discrepancies between generated outputs and reference charts. \textbf{(2) Fine-tuning the base model on our ReChartPrompt-240K dataset (``Base.+ReCha.'') significantly improves chart-to-code generation accuracy.} This improvement arises from the diverse, high-quality training data generated by conditioning on real-world chart prompts. Nonetheless, minor issues remain, such as slight mismatches in color or element positioning compared to the ground truth, indicating that supervised fine-tuning alone does not achieve perfect visual consistency. \textbf{(3) Incorporating the ChartSimRL algorithm further improves both visual and semantic alignment.} Notably, the model demonstrates enhanced color accuracy (as seen in the first column of Fig.~\ref{fig:dingxing}) and more faithful reproduction of arrow styles in the second column, reflecting improved attention to key factual details. \textbf{(4) ChartMaster competes favorably with GPT-4o.} Notably, the ChartMaster-7B model can generate charts that more closely resemble the ground truth than those from GPT-4o, especially excelling in ``mimicking'' chart attributes. Additional generation results in Appendix Fig. \ref{fig:dingxing2} consistently support these conclusions.

\section{Conclusion}
In this paper, we propose \textbf{ChartMaster}, a novel chart-to-code generation framework paired with a tailored reinforcement learning algorithm. By introducing \textbf{ReChartPrompt}, we address data homogeneity issues in prior work and build a highly diverse ReChartPrompt-240K dataset. Our \textbf{ChartSimRL} algorithm combines semantic and visual similarity rewards, enabling the model to generate chart code that closely matches original visuals. Experiments show ChartMaster achieves performance on par with GPT-4o in chart-to-code tasks. We will open source all resources to foster community development and advance research in this area.

Beyond its technical innovations, ChartMaster supports automated scientific reporting and empowers data-driven decision-making across a wide range of domains. While our current framework targets common chart types and Python-based code, expanding its scope to include a wider range of chart formats and programming languages is an exciting direction for future work.

\bibliography{iclr2026_conference}
\bibliographystyle{iclr2026_conference}
\clearpage
\appendix

\section{Implementation and Evaluation Details} \label{sec:details}
During the collection of arXiv papers, we explicitly exclude papers that are used as benchmarks to avoid potential data leakage. In the data generation stage, we apply a greedy sampling strategy to filter chart data, retaining only images of 12 predefined chart types and discarding all others. Then, we randomly select an instruction from $\mathbf{P}_{\text{rechart}}$ to prompt the Qwen2.5-VL-72B~\citep{bai2025qwen2} model to generate code via nucleus sampling, with a temperature of 0.1 and a top-p of 0.9.

For training, we use the Qwen2.5-VL-7B model~\citep{bai2025qwen2} in two stages. In Stage~1, we perform SFT on the entire ReChartPrompt-240K dataset with a learning rate of \(2 \times 10^{-5}\), batch size 128, and a cosine annealing scheduler for one epoch; the resulting model is saved for Stage~2. In Stage~2, ChartSimRL training is conducted on 10\% of the ReChartPrompt-240K dataset, using a smaller learning rate of \(5 \times 10^{-6}\) and generating \(M=4\) candidate codes per sample. Candidate sampling uses temperature 1.0, top-p 1.0, and top-k 80 to encourage diversity. The batch size remains 128 (32 samples \(\times\) 4 candidates each). 

For evalution, we assess the model's chart-to-code generation performance on multiple benchmarks.
\textbf{ChartMimic Direct Mimic Task} \citep{yang2024chartmimic}: This benchmark includes 600 chart images. GPT-4o scores (0–100) serve as high-level similarity metrics. Additionally, low-level F1 scores for text, layout, chart type, and color are computed from code execution for fine-grained analysis.
\textbf{Plot2Code Direct Asking} \citep{wu2024plot2code}: Metrics include code pass rate, text match rate, and a 10-point GPT-4V visual similarity score, jointly assessing code correctness and visual fidelity.
\textbf{ChartX Chart Redrawing Task} \citep{xia2024chartx}: This benchmark uses GPT-4 (0–5 scale) to evaluate code-generated chart redrawings.

\section{Standard Metrics} \label{sec:standard_metrics}
We consider two RGB images: the original chart image \( I_i \in \mathbb{R}^{H \times W \times 3} \) and the generated chart image \(\hat{I}_i \in \mathbb{R}^{H \times W \times 3}\), where \(H\) and \(W\) denote the height and width of the images respectively (both images are resized to the same height and width before comparison), and \(3\) corresponds to the RGB color channels. Below, we describe how to quantify the visual similarity between 
\( I_i \) and \( \hat{I}_i \) using metrics such as Mean Squared Error (MSE), Structural Similarity (SSIM) \citep{wang2004image}, and Peak Signal-to-Noise Ratio (PSNR) \citep{hore2010image}.
  
\subsection{Mean Squared Error}
The Mean Squared Error (MSE) is defined as:
\[
\mathrm{MSE}(I_i, \hat{I}_i) = \frac{1}{H \times W \times 3} \sum_{h=1}^{H} \sum_{w=1}^{W} \sum_{c=1}^{3} \left( I_i(h,w,c) - \hat{I}_i(h,w,c) \right)^2
\]
This formula computes the average squared difference between the pixel values of the two images over all spatial locations and color channels. A smaller MSE indicates higher similarity between \(I_i\) and \(\hat{I}_i\).

To convert the MSE into a similarity score, we define the MSE-based similarity as:
\[
\mathrm{MSE\_Similarity} = \frac{1}{1 + \mathrm{MSE}(I_i, \hat{I}_i)} \in (0,1]
\]
\begin{itemize}
    \item When \(\mathrm{MSE} \to 0\), \(\mathrm{MSE\_Similarity} \to 1\), indicating the images are almost identical.
    \item When \(\mathrm{MSE} \to \infty\), \(\mathrm{MSE\_Similarity} \to 0\), indicating large differences between the images.
\end{itemize}

\subsection{Structural Similarity}

The Structural Similarity (SSIM) is a perceptual metric that quantifies the similarity between two images by comparing local patterns of pixel intensities.  
It is computed on local sliding windows centered at each pixel location. For each window, local statistics including mean, variance, and covariance are calculated to evaluate the similarity.  
The final SSIM value for each channel is obtained by averaging these local SSIM values over all spatial positions, and the overall SSIM between two RGB images is computed by averaging over the three color channels.

Formally, for each color channel \(c \in \{R, G, B\}\), the SSIM is defined as:
\[
\mathrm{SSIM}_c (I_i^c, \hat{I}_i^c) = \frac{(2\mu_{I_i^c} \mu_{\hat{I}_i^c} + C_1)(2\sigma_{I_i^c \hat{I}_i^c} + C_2)}{(\mu_{I_i^c}^2 + \mu_{\hat{I}_i^c}^2 + C_1)(\sigma_{I_i^c}^2 + \sigma_{\hat{I}_i^c}^2 + C_2)}
\]
where
\begin{itemize}
  \item \(\mu_{I_i^c}\) and \(\mu_{\hat{I}_i^c}\) are the local means computed within the sliding window.
  \item \(\sigma_{I_i^c}^2\) and \(\sigma_{\hat{I}_i^c}^2\) are the local variances.
  \item \(\sigma_{I_i^c \hat{I}_i^c}\) is the local covariance.
  \item \(C_1 = (K_1 L)^2\) and \(C_2 = (K_2 L)^2\) are constants to stabilize the division, with default values \(K_1=0.01\), \(K_2=0.03\). \(L\) is the dynamic range of the pixel values. For 8-bit grayscale images, \(L=255\). In our implementation, all images are converted to \texttt{np.float32} and normalized by dividing by 255, so the pixel values are in the range \([0, 1]\). Therefore, \(L=1.0\) is used for SSIM calculation.

\end{itemize}

The overall mean SSIM between the two RGB images is then calculated by averaging over all spatial positions \((x,y)\) in each channel and then over the three channels:
\[
\mathrm{SSIM}(I_i, \hat{I}_i) = \frac{1}{3} \sum_{c=1}^3 \frac{1}{H \times W} \sum_{x=1}^H \sum_{y=1}^W \mathrm{SSIM}_c (I_i^c(x,y), \hat{I}_i^c(x,y)) \in [0,1]
\]
\begin{itemize}
    \item When $\mathrm{SSIM} \to 1$, the images are structurally almost identical.
    \item When $\mathrm{SSIM} \to 0$, there are significant structural differences between the images.
\end{itemize}

\subsection{Peak Signal-to-Noise Ratio}

Peak Signal-to-Noise Ratio (PSNR) is a widely used metric to measure the quality of reconstructed images compared to the original images. It is defined as:
\[
\mathrm{PSNR}(I_i, \hat{I}_i) = 10 \log_{10} \left( \frac{L^2}{\mathrm{MSE}(I_i, \hat{I}_i)} \right)
\]
where \(L\) is the dynamic range of the pixel values. For normalized images in \([0,1]\), \(L=1.0\).

In practical scenarios, PSNR values typically range in tens of decibels and can vary widely, which may cause instability during optimization. To mitigate this effect, we normalize the PSNR values within each rollout batch by dividing them by the maximum PSNR value in that batch:
\[
\mathrm{PSNR}_{\mathrm{norm}}(I_i, \hat{I}_i) = \frac{\mathrm{PSNR}(I_i, \hat{I}_i)}{\max_{\hat{I}_j \in \text{rollout batch}} \mathrm{PSNR}(I_j, \hat{I}_j)} \in (0,1]
\]
\begin{itemize}
    \item When \(\mathrm{PSNR}_{\mathrm{norm}} \to 1\), the reconstructed image \(\hat{I}_i\) is very similar to the original image \(I_i\).
    \item When \(\mathrm{PSNR}_{\mathrm{norm}} \to 0\), there exist significant differences between the images.
\end{itemize}

\begin{figure*}[h]
    \centering
    
    \begin{subfigure}[t]{1.0\textwidth}
        \centering
        \includegraphics[width=\linewidth]{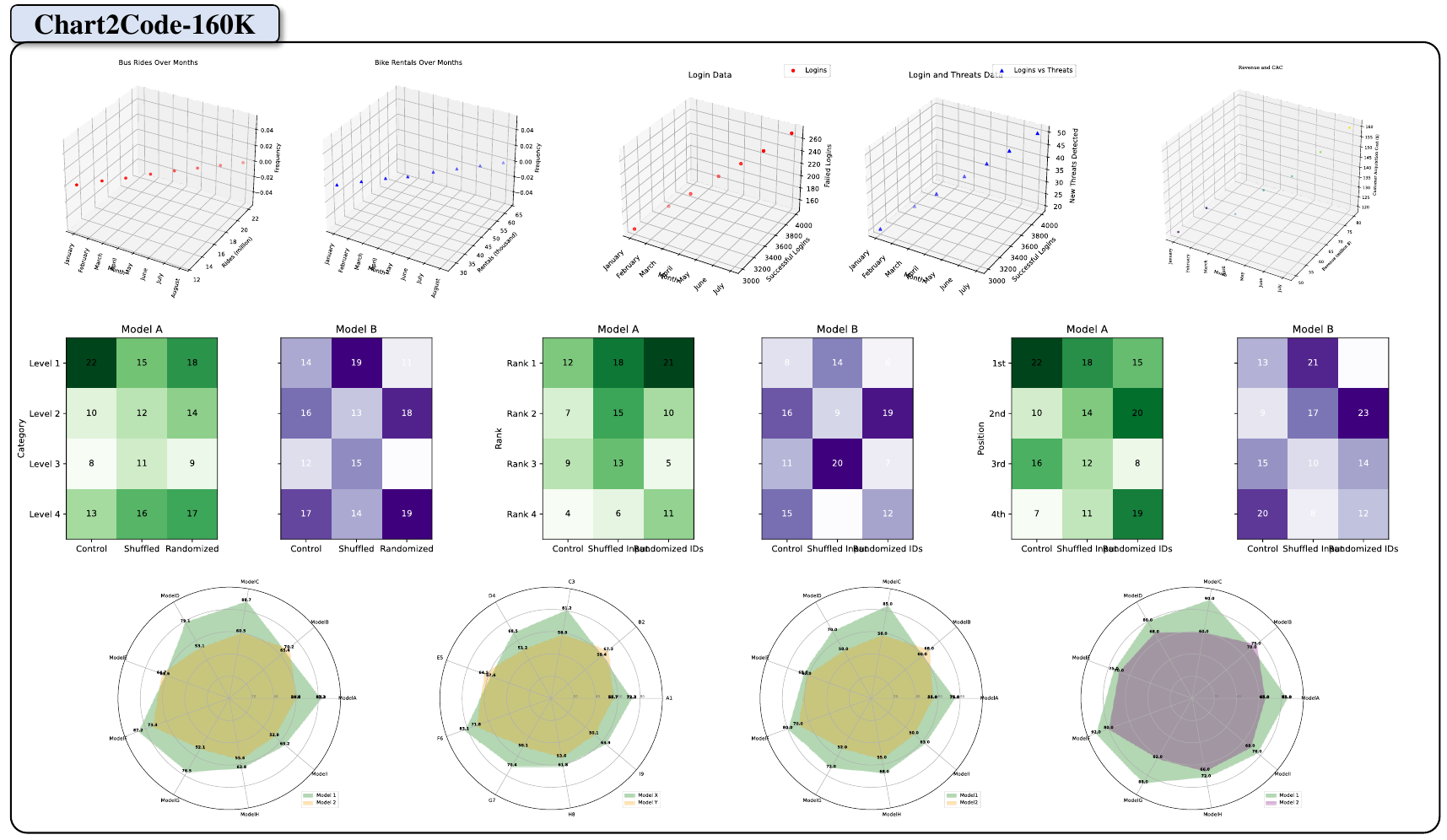} 
        \label{fig:image1}
    \end{subfigure}
    
    \vspace{0.5cm} 

    \begin{subfigure}[t]{1.0\textwidth}
        \centering
        \includegraphics[width=\linewidth]{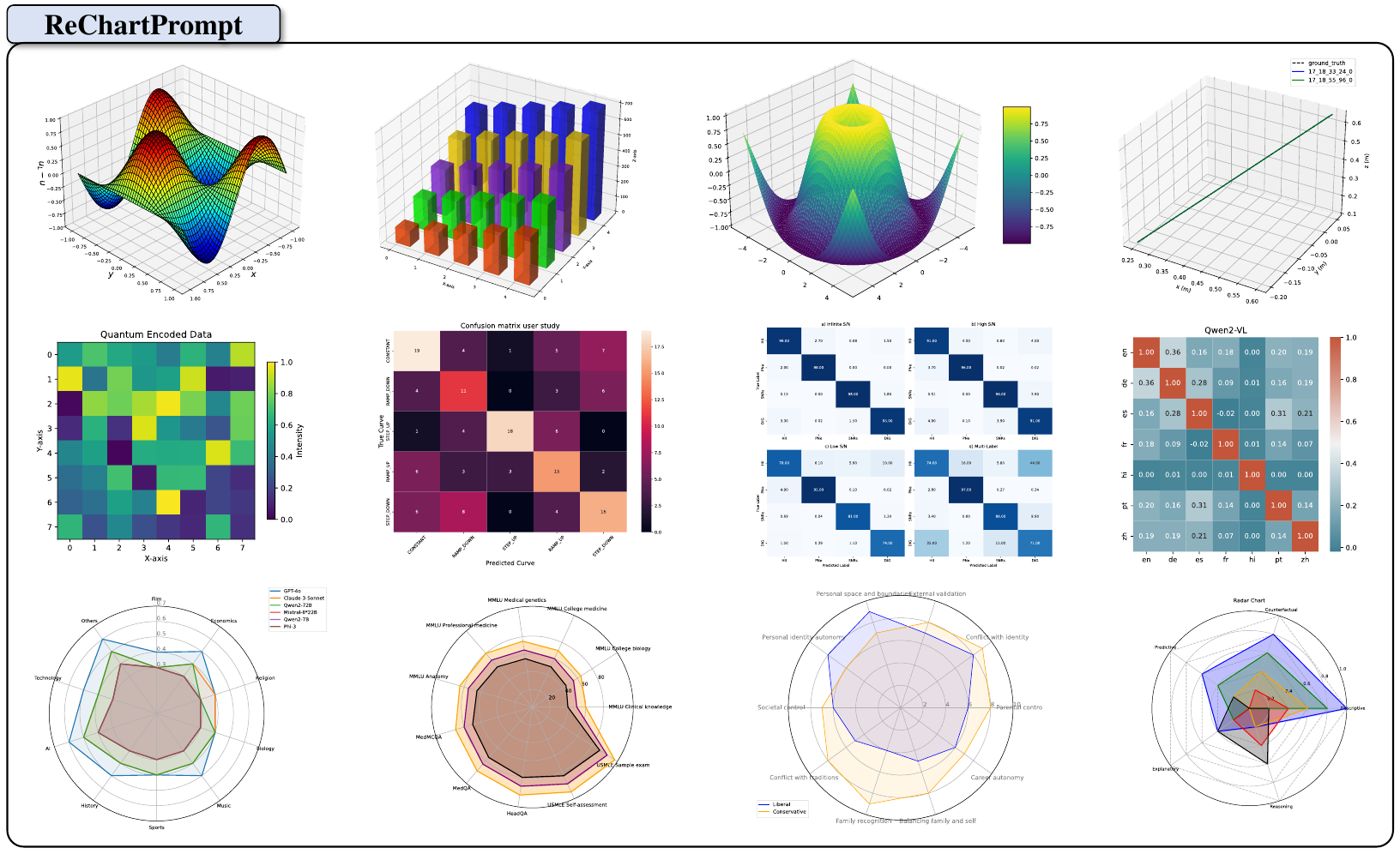} 
        \label{fig:image2}
    \end{subfigure}
    
    \caption{Dataset visualization. The charts in Chart2Code-160K exhibit homogenization, which affects diversity; the charts in ReChartPrompt demonstrate greater variety, especially in terms of textual content within the tables and layout attributes.} 
    \label{fig:dataset_visual}
\end{figure*}

\begin{figure}[t]
\centering
\setlength{\abovecaptionskip}{2pt}
\includegraphics[width=1.0\columnwidth]{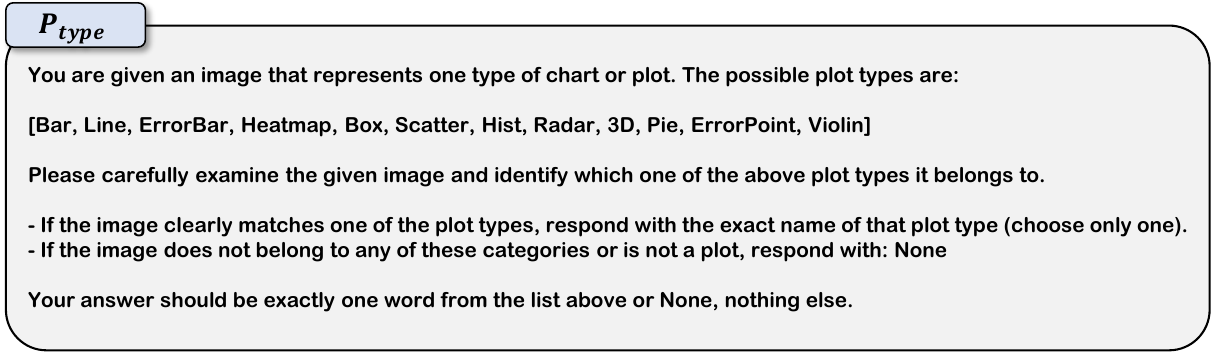} 
\caption{Prompt used for chart type classification (\(P_{\text{type}}\)). The Qwen2.5-VL-72B model is prompted with this template to assign each image to one of 12 predefined chart categories.
}
\label{fig:prompt_type}
\end{figure}

\begin{figure}[t]
\centering
\setlength{\abovecaptionskip}{2pt}
\includegraphics[width=1.0\columnwidth]{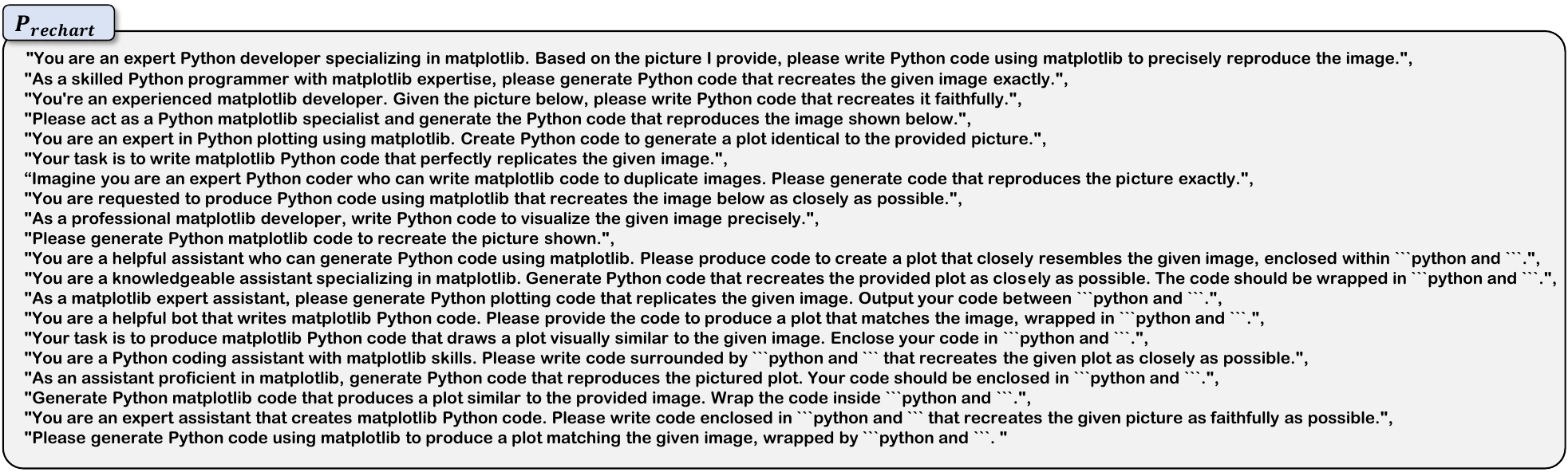} 
\caption{Prompt for chart-to-code generation (\(\mathbf{P}_{\text{rechart}}\)). Twenty diverse prompts are designed to instruct the Qwen2.5-VL-72B model to generate Python matplotlib code from chart images, enhancing instruction diversity.
}
\label{fig:prompt_rechart}
\end{figure}

\begin{figure}[t]
\centering
\setlength{\abovecaptionskip}{2pt}
\includegraphics[width=1.0\columnwidth]{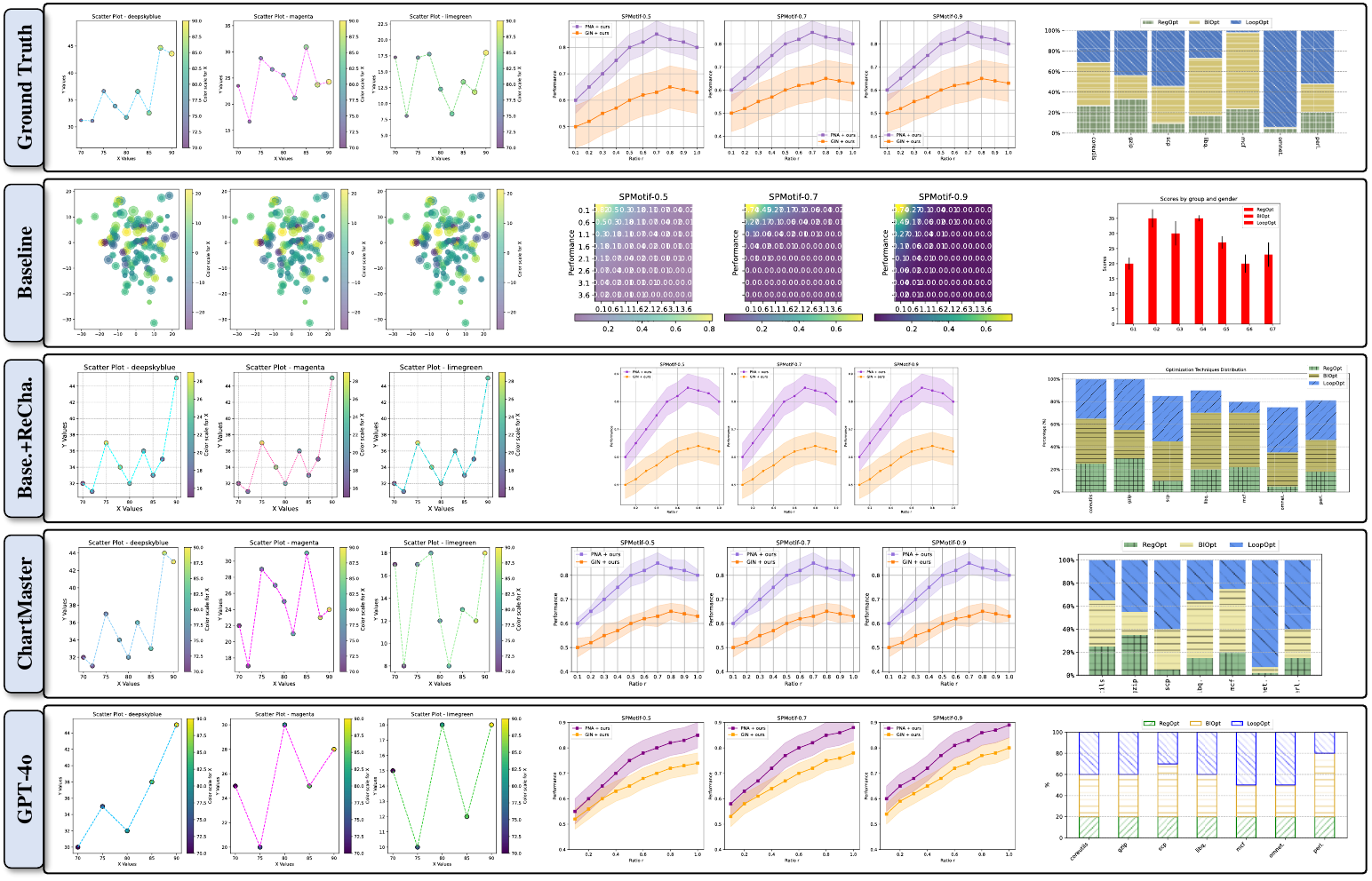} 
\caption{More test results of various models on the ChartMimic benchmark.
}
\label{fig:dingxing2}
\end{figure}

\end{document}